\documentclass[runningheads]{llncs}
\usepackage{graphicx}
\usepackage{amsmath,amssymb} 
\usepackage{color}

\usepackage{enumitem}
\newcommand{\vect}[1]{\boldsymbol{#1}}
\usepackage{amsmath,amssymb}
\DeclareMathOperator{\E}{\mathbb{E}}
\newcommand{\qref}[1]{Eq. \eqref{#1}}
\newcommand{\tref}[1]{Table \ref{#1}}
\newcommand{\fref}[1]{Fig. \ref{#1}}
\usepackage{textcomp}
\usepackage{algorithm}
\usepackage{algorithmicx}
\usepackage{algpseudocode}
\usepackage{graphicx}
\usepackage{subfig}
\usepackage{multirow}
\usepackage{cite}
\usepackage{caption}

\usepackage{pgfplots}
\usepackage[stable]{footmisc}

\makeatletter
\def\onedot{. }
\def\eg{\emph{e.g}\onedot} 
\def\ie{\emph{i.e}\onedot}

\def\etal{\emph{et al}\onedot}
\makeatother

\begin{document}
\pagestyle{headings}
\mainmatter

\title{Adversarial Open-World Person Re-Identification} 

\titlerunning{Adversarial Open-World Person Re-Identification}

\author{Xiang Li\inst{1},
Ancong Wu\inst{1} ,
and Wei-Shi Zheng\inst{1,2,3}\thanks{corresponding author}\orcidID{0000-0001-8327-0003}}

\authorrunning{X. Li,
A. Wu
and W.-S. Zheng}


\institute{Sun Yat-sen University\\
	\email{ \{lixiang47,wuancong\}@mail2.sysu.edu.cn \\
  wszheng@ieee.org}
\and
Inception Institute of Artificial Intelligence, United Arab Emirates
\and
%
Key Laboratory of Machine Intelligence and Advanced Computing, MOE
}

\maketitle

\begin{abstract}

In a typical real-world application of re-id, a watch-list (gallery set) of a handful of target people (\eg suspects) to track around a large volume of non-target people are demanded across camera views, and this is called the open-world person re-id. Different from conventional (closed-world) person re-id,
a large portion of probe samples are not from target people in the open-world setting. And, it always happens that a non-target person would look similar to a target one and therefore would seriously challenge a re-id system.
In this work, we introduce a deep open-world group-based person re-id model based on adversarial learning to alleviate the attack problem caused by similar non-target people.
The main idea is learning to attack feature extractor on the target people by using GAN to generate very target-like images (imposters), and in the meantime the model will make the feature extractor learn to tolerate the attack by discriminative learning so as to realize group-based verification.
The framework we proposed is called the adversarial open-world person re-identification, and this is realized by our Adversarial PersonNet (APN) that jointly learns a generator, a person discriminator, a target discriminator and a feature extractor, where the feature extractor and target discriminator share the same weights so as to makes the feature extractor learn to tolerate the attack by imposters for better group-based verification.
While open-world person re-id is challenging, we show for the first time that the adversarial-based approach helps stabilize person re-id system under imposter attack more effectively.
\end{abstract}

\section{Introduction}

\begin{figure}[t]
    \begin{center}
        \includegraphics[width=1\linewidth]{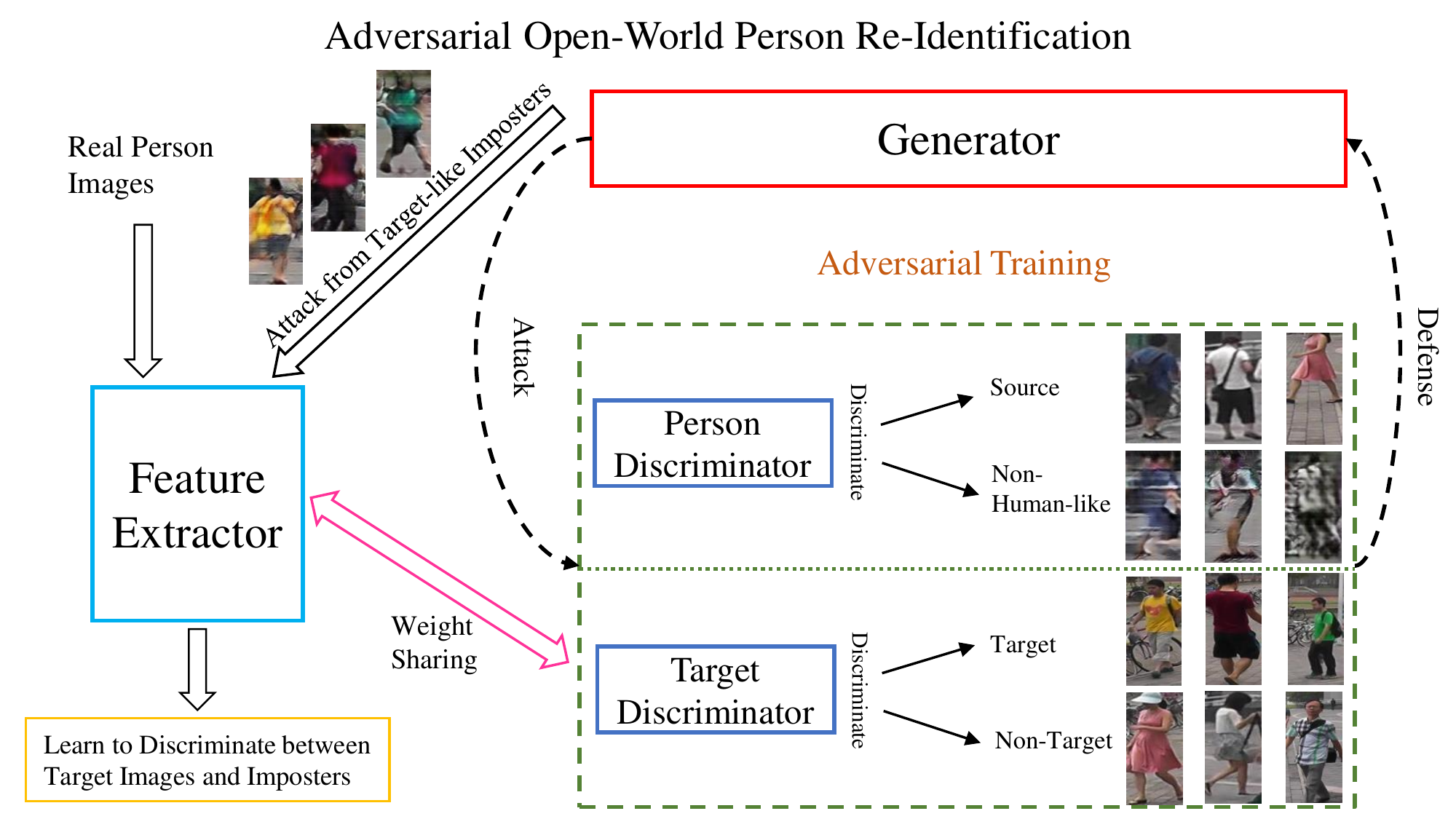}
    \end{center}

    \caption{Overview of adversarial open-world person re-identification.
    The goal for the generator is to generate target-like images, while we have two discriminators here. The person discriminator
    is to discriminate whether the generated images are from source dataset (\ie being human-like). And the target discriminator is to
    discriminate whether the generated images are of target people.
    By the adversarial learning, we aim to generate images beneficial for training a better feature extractor for telling target person images apart from non-target ones.}
    \label{fig:intro}
\end{figure}

Person re-identification (re-id), which is to match a pedestrian
across disjoint camera views in diverse scenes, is practical and useful for many fields, such
as public security applications and has gained increasing interests in recent
years
\cite{journals/ivc/Bedagkar-GalaS14,journals/pami/ZhengGX16,series/acvpr/978-1-4471-6295-7,journals/corr/ChenZZL17,journals/corr/WuCLWYZ16,conf/cvpr/LiaoHZL15,conf/cvpr/FarenzenaBPMC10,conf/cvpr/ZhengGX11,journals/corr/WuSH16,journals/corr/ZhuZLLCZ17,conf/bmvc/CancelaHG14,conf/icip/WangZXG16}. Rather than re-identifying every person in a multiple camera network, a typical real-world application is to re-identify or track only a handful of target people on a watch list (gallery set), which is called the open-world person re-id problem \cite{journals/pami/ZhengGX16,conf/bmvc/CancelaHG14,8011476}.
While target people will reappear in the camera network at different views, a large volume of non-target people, some of which could be very similar to target people, would appear as well. This contradicts to the conventional closed-world person re-id setting that all probe queries are belonging to target people on the watch list. In comparison, the open-world person re-id 
is extremely challenging because both target and non-target (irrelevant)
people are included in the probe set.

However, the majority of current person re-identification models are designed for the closed-world setting
\cite{journals/corr/ZhengZY17,journals/corr/WuCLWYZ16,journals/corr/ChenZZL17,journals/tcsv/TaoJWYL13,conf/cvpr/ZhengGX11,journals/corr/WuSH16,journals/corr/ZhuZLLCZ17,conf/cvpr/XiaoLOW16}
rather than the open-world one.
Without consideration of discriminating target and non-target people during learning, these approaches are not stable and could
often fail to reject a query image whose identity is not included in the gallery set. Zheng \etal \cite{journals/pami/ZhengGX16} considered this problem and proposed open-world group-based verification model. Their model is based on hand-crafted feature and transfer-learning-based metric learning with auxiliary data, but the results are still far from solving this challenge. More importantly, the optimal feature representation and target-person-specific information for open-world setting have not been learned. 

In this work, we present an adversarial open-world person re-identification framework for
1) learning features that are suitable for open-world person re-id, and
2) learning to attack the feature extractor by generating very target-like imposters and make person re-id system learn to tolerate it for better verification.
An end-to-end deep neural network is designed to realize the above two objectives, and an overview of this pipeline is shown in \fref{fig:intro}.
The feature learning and the adversarial learning are mutually related and learned jointly, meanwhile the generator and the feature extractor are learned from each other iteratively to enhance both the efficiency of generated images and the discriminability of the feature extractor.
To use the unlabeled images generated, we further incorporate a label smoothing regularization for imposters (LSRI)
for this adversarial learning process. LSRI allocates equal probabilities of being any non-target people and zero probabilities of being target people to the generated target-like 
imposters, and it would  further improve the discrimination ability of the feature extractor for distinguishing real target people from fake ones (imposters). 

While GAN has been attempted in Person re-id models recently in \cite{journals/corr/ZhengZY17,zhong2018camera,deng2018image} for generating images adapted from source dataset so as to enrich the training dataset on target task. 
However, our objective is beyond this conventional usage.
By sharing the weights between feature extractor and target discriminator (see Fig. \ref{fig:APN}), our adversarial learning makes 
the generator and feature extractor interact with each other in an end-to-end framework. This interaction not only makes the generator
produce imposters
look like target people, but also more importantly makes the feature extractor learn to tolerate the attack by imposters for better group-based verification.

In summary, our contributions are more on solving the open-world challenge in person re-identification. It is the first time to formulate the open-world group-based person re-identification under an adversarial learning framework. By learning to attack and learning to defend, we realize four progresses in a unified framework, including generating very target-like imposters, mimicking imposter attacking, discriminating imposters from target images and learning re-id feature to represent. Our investigation suggests that adversarial learning is a more effective way for stabilizing person re-id system undergoing imposters.

\section{Related Work}

\begin{description}[style=unboxed,leftmargin=0cm]
    \item[Person Re-Identification:] Since person re-identification targets to identify
        different people, better feature representations
        are studied by a great deal of recent research. Some of the research try to seek more discriminative/reliable hand-crafted features        \cite{conf/cvpr/FarenzenaBPMC10,journals/pami/KviatkovskyAR13,ma2012bicov,DBLP:journals/ivc/MaSJ14,conf/cvpr/LiaoHZL15,journals/corr/WuCLWYZ16,journals/pr/LiuGL14}. Except that, learning the best matching metric        \cite{conf/eccv/HirzerRKB12,conf/cvpr/KostingerHWRB12,conf/cvpr/MignonJ12,journals/tcsv/TaoJWYL13,conf/cvpr/ZhengGX11,journals/tcsv/ChenZLY17,journals/corr/ChenZZL17}
        is also widely studied for solving the cross-view change in different environments.
        With the rapid development of deep learning, learning to represent from images    \cite{conf/cvpr/AhmedJM15,conf/cvpr/ChengGZWZ16,journals/corr/DingLWC15,conf/cvpr/LiZXW14}
        is attracted for person re-id, and in particular Xiao \etal
        \cite{conf/cvpr/XiaoLOW16}
        came up with domain guided drop out model for training CNN with multiple domains so as to improve the feature
        learning procedure. Also, recent deep approaches in
        person re-identification
        are found to unify feature learning and metric learning \cite{journals/corr/WuSH16,journals/corr/ZhuZLLCZ17,conf/cvpr/AhmedJM15,conf/nips/SubramaniamCM16}.
        Although these deep learning methods are expressive for large-scale datasets, they tend to be resistless
        for noises and incapable of distinguishing non-target people apart from the target ones, and thus becomes unsuitable for the open-world setting.
        In comparison, our deep model aims to model the effect of non-target people during training and optimize the person re-id in the open-world setting.

    \item[Towards Open-World Person Re-Identification:] Although the majority of works on person re-id are focusing on the closed-world setting, a few works have been reported on addressing the open-world setting.
    The work of Candela \etal
        \cite{conf/bmvc/CancelaHG14}
        is based on Conditional Random Field (CRF) inference attempting
        to build connections between cameras towards open-world person re-identification.
        But their work lacks the ability to distinguish very similar identities, and with some deep CNN models coming up, features from multiple camera views
        can be well expressed by joint camera learning. Wang \etal
        \cite{conf/icip/WangZXG16}
        worked out an approach by proposing
        a new subspace learning model suitable for open-world scenario. However,
        group-based setting and interference defense is not considered. Also, their model requires a large volume of extra unlabeled data.
        Zhu \etal \cite{8011476} proposed a novel hashing method for fast search in the open-world setting.
        However, Zhu \etal aimed at large scale open-world re-identification
        and efficiency is considered primarily. Besides, noiseproof ability is not
        taken into account. The most correlated work with this paper is formulated
        by Zheng \etal
        \cite{journals/pami/ZhengGX16,conf/cvpr/ZhengGX12}, where the group-based verification towards open-world person re-identification was proposed. They came up with
        a transfer relative distance comparison model (t-LRDC), learning a
        distance metric and transferring non-target data to target data in order to overcome data sparsity.
        Different from the above works, we present the first end-to-end learning model to unify feature learning and verification modeling to address the open-world setting.
        Moreover, our work does not require extra auxiliary datasets to mimic attack of imposters, but integrates an adversarial processing to make re-id model learn to tolerate the attack.

     \item[Adversarial Learning:] In 2014, Szegedy \etal
        \cite{journals/corr/SzegedyZSBEGF13}
        have found out that tiny noises in samples can lead deep classifiers to mis-classify, even if these adversarial samples can be easily
        discriminated by human. Then many researchers have been working on adversarial training. Seyed-Mohsen \etal \cite{conf/cvpr/Moosavi-Dezfooli16}
        proposed DeepFool, using the gradient of an image to produce a minimal noise that fools deep
        networks. However, their adversarial samples are towards
        individuals and the relation between target and non-target groups is not modelled. Thus, it does not well fit into the group-based setting.
        Nicolas Papernot \etal \cite{journals/corr/PapernotMJFCS15}
        formulated a class of algorithms by using knowledge of deep neural networks (DNN)
        architecture for crafting adversarial samples. However, rather than forming a general
        algorithm for DNNs, our method is more specific for group-based person verification
        and the imposter samples generated are more effective to this scenario.
        Later, SafetyNet by Lu \etal
        \cite{journals/corr/LuIF17}
        was proposed with an RBF-SVM in full-connected layer to detect adversarial samples.
        However, we perform the adversarial learning at feature level to better attack the learned features.

\end{description}


\section{Adversarial PersonNet}

\subsection{Problem Statement}

In this work, we concentrate on open-world person re-id by group-based verification. The group-based verification is to ensure a re-id system to identify whether a query person image comes from target people on the watch list. In
this scenario, 
people out of this list/group are defined as 
non-target people.

Our objective is to unify feature learning by deep convolution networks and adversarial learning together so as to make
the extracted feature robust and resistant to noise for discriminating between target people and non-target ones. The adversarial learning is to generate target-like imposter images to
attack the feature extraction process and simultaneously make the whole model learn to distinguish these attacks.
For this purpose, we propose a novel deep learning model called Adversarial PersonNet (APN) that suits open-world person re-id.

To better express our work under this setting in the following sections, we suppose that $N_T$ target training images
constitute a target sample set $X_T$ sampled from $C_T$ target people.
Let $\vect{x}^T_i$ indicate the $i$th target image and $y^T_i \in Y_T$ represents the corresponding
person/class label. The label set $Y_T$ is denoted by $Y_T = \{ y^T_1 ,..., y^T_{N_T} \}$ and there are $C_T$
target classes in total. Similarly, we are given a set of $C_S$ non-target training classes containing $N_S$ images, denoted as $X_S = \{ \vect{x}^S_1 ,..., \vect{x}^S_{N_S} \}$,
where $\vect{x}^S_i \in X_S$ is the $i$th non-target image. $y^S_i$ is the class of $\vect{x}^S_i$
and $Y_S = \{ y^S_1 ,..., y^S_{N_S} \}$. Note that there is no identity overlap between target people and non-target people. Under
open-world setting, $N_S \gg N_T$. The problem is to better determine whether a person is on
the target-list; that is for a given image $\vect{x}$ without knowing its class $y$,
determine if $y \in Y_T$. We use $f(\vect{x},\vect{\theta})$ to
represent the extracted feature from image $\vect{x}$, and $\vect{\theta}$ is the weight of
the feature extraction part of the CNN.

\begin{figure*}[t]
    \begin{center}
        \includegraphics[width=\linewidth, height=0.40\linewidth]{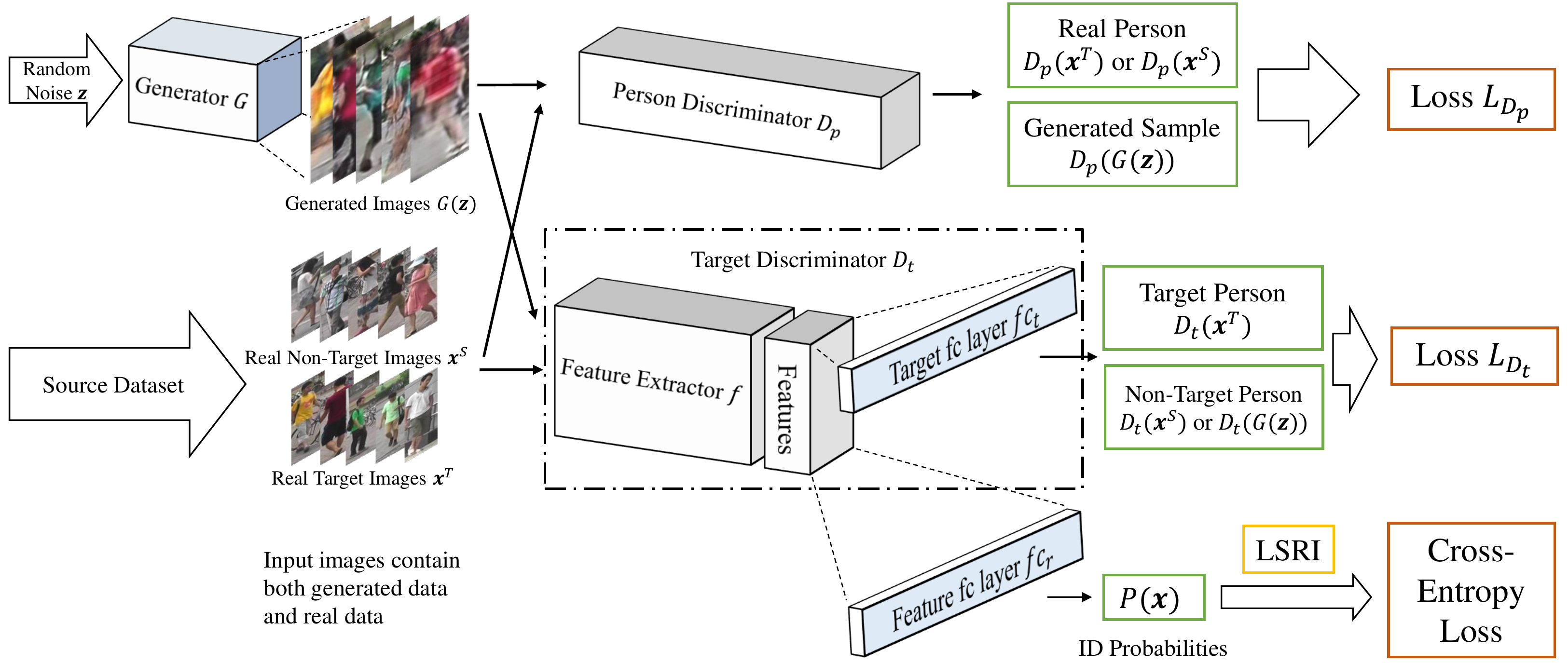}
    \end{center}
    \caption{Adversarial PersonNet structure. Two discriminators $D_p$ and $D_t$ accept samples
    from both datasets and generator $G$. 
    Since $D_t$ shares the same weights with
    feature extractor $f$, we represent them as the same cuboid in this figure.}
    \label{fig:APN}
\end{figure*}

\subsection{Learning to Attack by Adversarial Networks}

Always, GANs are designed to generate images similar to those in the \emph{source set}, which is constituted by both target and non-target image sets. A generator
$G$ and a discriminator $D_p$ are trained adversarially.
However the generator $G$
normally only generates images looking like the ones in the source set and the discriminator $D_p$ discriminates the generated images from the source ones. In
this case, our source datasets are all pedestrian images, so we call such $D_p$ the \emph{person
discriminator} in response to its ability of determining whether an image is of pedestrian-like images. 
$D_p$ is trained by minimizing the following loss function:
\begin{equation}
    \label{q:ldp}
    L_{D_p} = -\frac{1}{m}\sum_{i=1}^{m}[\log{D_p(\vect{x})}
    + \log{(1 - D_p(G(\vect{z})))}],
\end{equation}
where $m$ is the number of samples, $\vect{x}$ represents image from source dataset and $\vect{z}$ is a noise randomly generated.

Suppose that there is a pre-trained feature extractor for person re-id task and in an attempt to steer generator $G$ to produce not only pedestrian-like but also
feature attacking images towards this feature extractor, we design a paralleled discriminator $D_t$ with the following definition:
\begin{equation} \label{eq:Dt}
    D_t(\vect{x}) = fc_t(f(\vect{x}, \vect{\theta})).
\end{equation}
The discriminator $D_t$ is to determine whether an image will be regarded as target image
by feature extractor.
$f(\vect{x}, \vect{\theta})$ indicates that part of $D_t$ has the same network structure as feature extractor $f$ and shares the same weights $\vect{\theta}$ (Actually, the feature extractor can be regarded as a part of $D_t$.). $fc_t$ means a full-connected layer following the feature extractor apart from the one connected to original CNN (with a fc layer used to pre-train the feature extractor).
So $D_t$ shares the same ability of target person discrimination with
the feature extractor.
To induce the generator $G$ for producing target-like images for attacking and ensure the discriminator $D_t$ to tell the non-target and generated imposters apart from the target ones, we formulate a paralleled
adversarial training of $G$ and $D_t$ as
\begin{equation} \label{eq:GDt}
\begin{split}
    \min_{G}\max_{D_t}V_t(D_t, G) &= \E_{\vect{x}^T \sim X_T}[\log{D_t(\vect{x}^T)}] \\
    &+ \E_{\vect{x}^S \sim X_S}[\log{(1 - D_t(\vect{x}^S))}] \\
    &+ \E_{\vect{z} \sim p_{\vect{z}}(\vect{z})}[\log{(1 - D_t(G(\vect{z})))}].
\end{split}
\end{equation}
We train $D_t$ to maximize $D_t(\vect{x})$ when passed by a target image but minimize it when passed by a non-target image or a generated imposter image by $G$. Notice that this
process only trains the final $fc_t$ layer of $D_t$ without updating the feature
extractor weights $\vect{\theta}$, to prevent the feature extractor from being affected by discriminator learning when the generated images are not good enough. We call $D_t$ the \emph{target discriminator}. 
And we propose the loss function $L_{D_t}$ for the training process of target discriminator $D_t$:
\begin{equation}
     \label{q:ldt}
         \begin{cases}
             &L_{D_t} = -\frac{1}{m}\sum_{i=1}^{m}[\log{Q_t(\vect{x})}
             + \log{(1 - D_t(G(\vect{z})))}], \\
             &Q_t(\vect{x}) = \begin{cases}
                 D_t(\vect{x}), & \text{$\vect{x} \in X_T$}; \\
                 1-D_t(\vect{x}), & \text{$\vect{x} \in X_S$}.
             \end{cases}
         \end{cases}
     \end{equation}
We integrate the above into a standard GAN framework 
as
follows:
\begin{equation} \label{eq:APN}
\begin{split}
    &\min_{G}\max_{D_p}\max_{D_t}V'(D_p, D_t, G) = \\
    &\E_{\vect{x}^T \sim X_T}[\log{D_p(\vect{x}^T)} + \log{D_t(\vect{x}^T)}] \\
    &+ \E_{\vect{x}^S \sim X_S}[\log{D_p(\vect{x}^S)} + \log{(1 - D_t(\vect{x}^S))}] \\
    &+ \E_{\vect{z}\sim p_{\vect{z}}(\vect{z})}[\log{(1 - D_p(G(\vect{z})))} + \log{(1 - D_t(G(\vect{z})))}]. \\
\end{split}
\end{equation}

\begin{figure}[t]
  \begin{minipage}[c]{0.6\textwidth}
    \begin{center}
    \includegraphics[width=\textwidth]{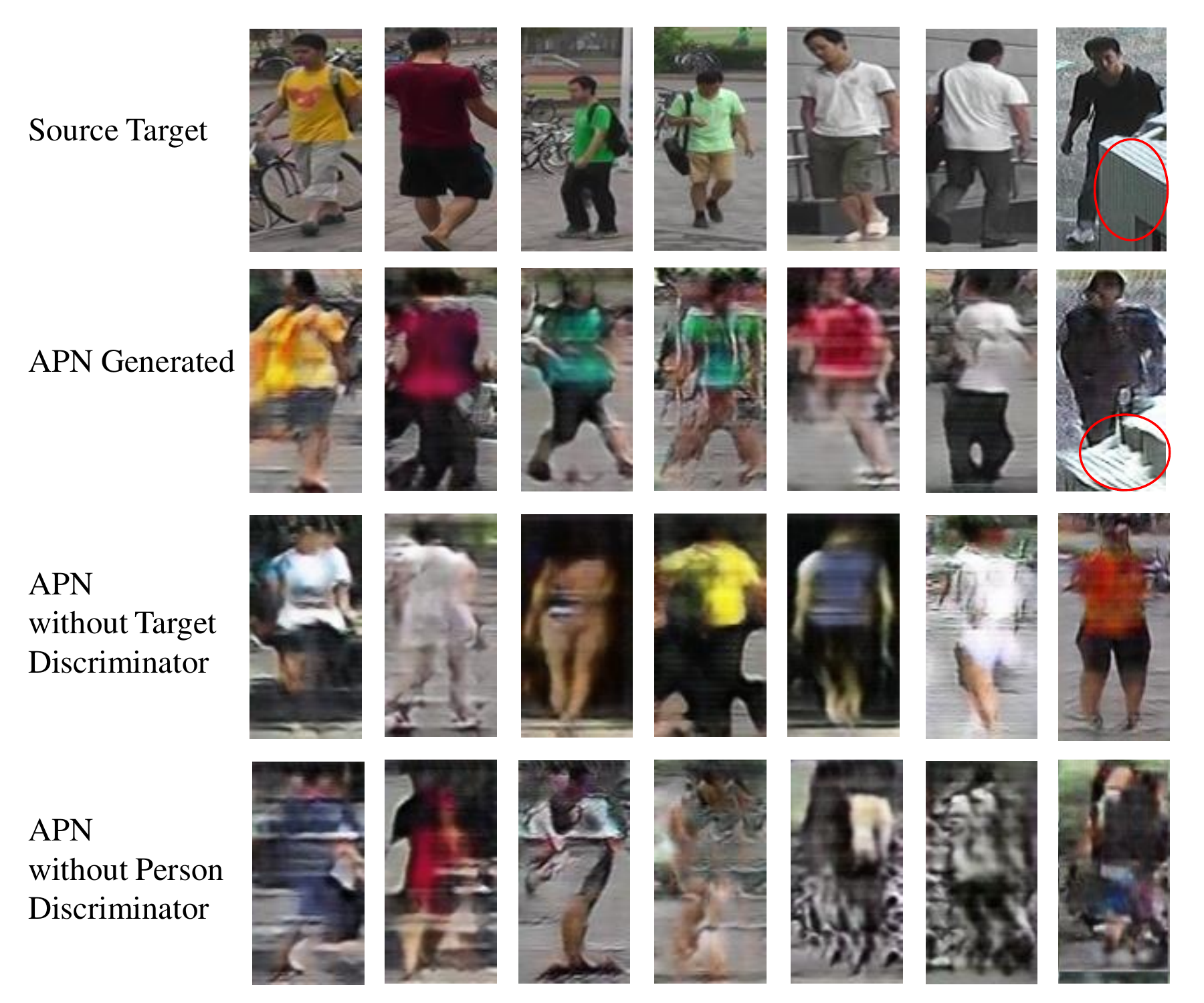}
    \end{center}
    \end{minipage}\hfill
    \begin{minipage}[c]{0.38\textwidth}
    \caption{Examples of generated images.
    Although images produced by the generator are based on random noises, we can tell that the imposters generated by APN are very similar to targets. These similarities are mostly based on clothes, colors and postures (\eg the fifth column). Moreover, surroundings are learned by APN as shown in the seventh column in the red circle.}
    \label{fig:samples}
    \end{minipage}
\end{figure}

The collaboration of generator and couple discriminators is illustrated in \fref{fig:APN}. While GAN
with only person discriminator will force the generator $G$ to produce source-like person images, with the incorporation of the loss of target discriminator $D_t$,
$G$ is more guided to produce very much target-like imposter images. The target-like imposters, generated based on the discriminating ability of feature extractor, satisfy the
usage of attacking the feature extractor.
Examples of images generated by APN are shown in \fref{fig:samples}
together with the target images and the images generated by controlled groups (APN without target discriminator $D_t$ and APN without person discriminator $D_p$) to indicate that our network indeed has the ability
to generate target-like images.
The generator $G$ is trained to fool the target discriminator in the feature space, so that the generated adversarial images can attack the re-id system.
While the target discriminator $D_t$ is mainly to tell these attack apart from the target people so as to defend the re-id system.

\subsection{Joint Learning of Feature Representation and Adversarial Modelling}

We finally aim to learn robust person features that are tolerant to imposter attack for open-world group-based person re-id.
For further utilizing the generated person images to enhance the performance,
we jointly learn feature representation and adversarial modelling in a semi-supervised way.

Although the generated images look similar
to target images, they are regarded as imposter samples, and we wish to incorporate unlabeled generated imposter samples. Inspired by the smoothing regularization \cite{journals/corr/ZhengZY17}, we modify the LSRO \cite{journals/corr/ZhengZY17} in order to make it more suitable for group-based verification by setting the probability of an unlabeled generated imposter sample $G(\vect{z})$ belonging to an existing known class $k$ as follows:
\begin{equation} \label{eq:LSRI}
    q_{LSRI}(k)(G(\vect{z})) = \begin{cases}
        \frac{1}{C_S}, & \text{$k \in Y_S$}; \\
        0, & \text{$k \in Y_T$},
    \end{cases}
\end{equation}
Compared to LSRO, we do not allocate a uniform distribution on each unlabeled data sample over all classes (including both target and non-target ones), but only allocate a uniform distribution on non-target classes. This is significant because we attempt to separate imposter samples from target classes.
The modification is exactly for the defense towards the attack of imposter samples.
By using this regularization, the generated imposters are more trending to be far away from
target classes and have equal chances of being non-target. We call the modified regularization in \qref{eq:LSRI} as label smoothing regularization for imposters (LSRI).

Hence for each input sample $\vect{x}_i$, we set its ground truth class distribution as:
\begin{equation} \label{eq:qk}
    q(k) = \begin{cases}
        1, & \text{$k = y_i$ and $\vect{x}_i \in X_T \cup X_S$}; \\
        0, & \text{$k \neq y_i$ and $\vect{x}_i \in X_T \cup X_S$, or $\vect{x}_i \in X_G$ and $k \in Y_T$}; \\
        \frac{1}{C_S}, & \text{$\vect{x}_i \in X_G$, and $k \in Y_S$}; \\
    \end{cases}
\end{equation}
where $y_i$ is the corresponding label of $x_i$, and we let $\vect{x}^G_i$ be the $i$th generated image and denote $X_G = \{ \vect{x}^G_1 ,..., \vect{x}^G_{N_G} \}$ as the set of
generated imposter samples.
With \qref{eq:qk}, we can now learn together with our feature extractor (\ie weights $\vect{\theta}$).
By such a joint learning, the feature learning part will become more discriminative between target and
target-like imposter images.

\subsection{Network Structure}

We now detail the network structure.
As shown in \fref{fig:APN},
our network consists of two parts: 1) learning robust feature representation, and 2) learning to attack by adversarial networks.
For the first part, we train the feature extractor from source datasets and generated attacking samples.
In this part, features are trained to be robust and resistant to imposter samples. LSRI is applied in this part to
differentiate imposters from target people. 
Here, a full-connected layer $fc_r$ is connected to feature extractor $f$ at this stage, and we call it the \textit{feature fc layer}.
For the second part, as shown in
\fref{fig:APN}, our learning attack by adversarial networks is a modification of DCGAN \cite{journals/corr/WuSH16}.
We combine modified DCGAN with couple discriminators
to form an adversarial
network. The generator $G$ here is modified to produce target-like imposters specifically as an
attacker. And the target discriminator $D_t$ defends as discriminating target from non-target people.
Of course, in this discriminator, a new $fc$ layer
is attached to the tail of feature extractor $f$, and we mark it $fc_t$, also called \textit{target fc layer}, used
to discriminate target from non-target images at the process of learning to attack by adversarial networks. By \qref{eq:Dt}, $D_t$ is the combination of $f$ and target fc layer $fc_t$.

\section{Experiments\footnote{We correct a bug in our code and conduct a better finetune using new implementations in PyTorch for all deep learning methods including ours and the compared ones. Our model is still effective and overall the best. Please refer to the new results.}}

\subsection{Group-based Verification Setting}

We followed the criterion defined in \cite{journals/pami/ZhengGX16} for evaluation of open-world group-based person re-id.
The performance of how well a true target can be verified correctly and
how badly a false target can be verified as true incorrectly is indicated by true target rate (TTR) and false target rate (FTR), which are defined as follows:
\begin{equation}\scriptsize
    \begin{aligned}
        \textbf{True Target Rate(TTR)} =\#TTQ / \#TQ, \ 
        \textbf{False Target Rate(FTR)} = \#FNTQ / \#NTQ,
    \end{aligned}
\end{equation}
where $TQ $ is the set of query target images from target people, $NTQ$ is the set of query non-target images from non-target people, $TTQ$ is the set of query target images that are verified as target people, and $FNTQ$ is the set of query non-target images that are verified as target people.

To obtain TTR and FTR, we follow the two steps below: 1) For each target person, there is a set of images $S$ (single-shot or multi-shot) in gallery set. Given a query sample $\vect{x}$, the distance between sample $\vect{x}$ and a set $S$ is the minimal distance between that  sample and any target sample of that set; 2) Whether a query image is verified as a target person is determined by comparing the distance to a threshold $r$. By changing the threshold $r$, a set of TTR and FTR values can be obtained. A higher TTR value is preferred when FTR is small.

In our experiments, we conducted two kinds of verification as defined in \cite{journals/pami/ZhengGX16}， namely Set Verification (\ie whether a query is one of the persons in the target set, where the target set contains all target people.) and Individual Verification (\ie whether a query is the true target person. For each target query image, the target set contains only this target person). In comparison, Set Verification is more difficult. Although determining whether a person image belongs to a group of target people seems easier, it also gives more chances for imposters to cheat the classifier, producing more false matchings \cite{journals/pami/ZhengGX16}. Also note that it is more important to compare the TTR values when FTR is low since it is expected to have more true matching when the false matching is less.

\subsection{Datasets \& Settings}

We evaluated our method on three datasets including Market-1501
\cite{zheng2015scalable},
CUHK01
\cite{li2012human},
and CUHK03
\cite{li2014deepreid}.
For each dataset, we randomly selected 1\% people as target people and the
rest as non-target. Similar to \cite{journals/pami/ZhengGX16}, for target people, we separated images of each target people into
training and testing sets by half.
Since only four images are available in CUHK01, we chose one for training, two for gallery (reduce to one in single-shot case) and one for probe.
Our division guaranteed that probe and gallery images are from diverse cameras for each person.
For non-target people, they were divided into training and testing sets by half in person/class level to ensure there is no overlap on identity.
In testing phase, two images of each target person in testing set were randomly selected to form gallery set, and the remaining images were selected to form query set.
In the default setting, all images of non-target people in testing set were selected to form query set.
The data split was kept the same for all evaluations on our and the compared methods.
Specifically, the data split is summarized below:
\begin{description}[style=unboxed,leftmargin=0cm]
    \item[CUHK01] CUHK01 contains 3,884 images of 971 identities from two camera views. In our experiment, 9 people were marked
        as target and 1,888
        images of 472 people were selected to to form the non-target training set. The testing
        set of non-target people contains 1,960 images of 490 people.
    \item[CUHK03] CUHK03 is larger than CUHK01 and some images were
        automatically detected. A total of 1,467 identities were divided into 14 target
        people, 712 training non-target people and 741 testing non-target people. The numbers of
        training and testing non-target images were 6,829 and 7,129 respectively.
     \item[Market-1501] Market-1501 is a large-scale dataset containing a total of
        32,668 images of 1,501 identities.
        We randomly selected 15 people as target and 728 people as non-target
        to form the training set containing a total of 17,316 images, and the testing non-target set
        contains 758 identities with 14,573 images.
\end{description}

Under the above settings, we evaluated our model together with selected popular re-id models.
We used the traditional hand-crafted features suggested in their original papers\footnote{We applied these metric learning methods to features extracted by ResNet-50 in the previous version to show our improvement on ResNet-50, but we found that without combining the whole process as an end-to-end learning, the performance of ResNet-50 feature based metric learning methods will largely decrease.} while evaluating metric learning methods such as t-LRDC \cite{journals/pami/ZhengGX16}, XICE \cite{8011476}, XQDA \cite{conf/cvpr/LiaoHZL15} and
CRAFT \cite{journals/corr/ChenZZL17}.

\subsection{Implementation Details}

In our APN, we used ResNet-50
\cite{conf/cvpr/HeZRS16}
as the feature extractor in the target discriminator. The generator and person discriminator is based on DCGAN
\cite{journals/corr/RadfordMC15}.
At the first step of our procedure, we pre-trained the feature extractor using auxiliary datasets, 3DPeS
\cite{baltieri2011_308},
iLIDS
\cite{journals/corr/WangGZW16},
PRID2011
\cite{conf/scia/HirzerBRB11}
and Shinpuhkan
\cite{Kawanishi_shinpuhkan2014:a}.
These datasets were only used in the pre-training stage for the feature extractor.
In pre-training, we used stochastic gradient descent with momentum 0.9. The learning rate
was 0.1 at the beginning and multiplied by 0.1 every 10 epochs.
Then, the adversarial part of APN was trained using ADAM optimizer \cite{journals/corr/KingmaB14} with parameters $\beta_1 = 0.5$ and $\beta_2 = 0.99$.
Using the target dataset for evaluation, the person discriminator $D_p$ and generator $G$ were pre-trained for 20 epochs. Then, the target discriminator $D_t$ together with the person discriminator $D_p$, and the generator $G$ were
trained jointly for $k_1=15$ epochs, where $G$ is optimized twice in each iteration to prevent losses of discriminators from going to zero.
Finally, the feature extractor was trained again for $k_2=20$ epochs with
a lower learning rate starting from 0.001 and multiplied by 0.1 every
10 epochs. The above procedure was executed repeatedly as an adversarial process.

\begin{table}[t]
\centering
\caption{Comparison with typical person re-identification: TTR (\%) against FTR}
\label{t:state-of-art}
\resizebox{\textwidth}{!}{
\begin{tabular}{c|c c c c c c|c c c c c c|c c c c c c}
\hline
Dataset & \multicolumn{6}{c|}{Market-1501} & \multicolumn{6}{c|}{CHUK01} & \multicolumn{6}{c}{CUHK03} \\ \hline
FTR         & 0.1\%          & 1\%   & 5\%   & 10\%  & 20\%  & 30\%   & 0.1\%   & 1\%   & 5\%   & 10\%  & 20\%  & 30\% & 0.1\%   & 1\%   & 5\%   & 10\%  & 20\%  & 30\%   \\ \hline
Evaluation    & \multicolumn{18}{c}{Set Verification}    \\ \hline
t-LRDC \cite{journals/pami/ZhengGX16} & 3.00  & 18.88  & 42.06 & 51.07 & 65.24 & 75.54 & 5.56 & 5.56 & 38.89 & 50.00 & 66.67 & 83.33 & 8.87 & 20.16 & 32.66 & 37.90 & 49.60 & 58.06 \\
XICE \cite{8011476} & 6.77 & 21.69 & 45.17 & 58.88 & 73.68 & 81.80 & 11.11 & 33.33 & 44.44 & 55.56 & 72.22 & 83.33 & 3.14 & 13.03 & 31.65 & 44.36 & 59.70 & 69.98 \\
GOG+XQDA \cite{conf/cvpr/LiaoHZL15} & 0.43 & 2.15 & 9.01 & 17.17 & 28.76 & 39.06 & 0 & 5.56 & 33.33 & 38.89 & 66.67 & 72.22 & 10.48 & 16.53 & 27.02 & 36.69 & 53.63 & 60.48 \\
LOMO+XQDA \cite{conf/cvpr/LiaoHZL15} & 4.72 & 14.16 & 35.62 & 46.35 & 58.80 & 65.67 & 0 & 5.56 & 38.89 & 44.44 & 72.22 & \textbf{88.89} & 25.81 & 38.31 & 51.61 & 61.69 & 71.77 & 83.47 \\
hiphop+CRAFT \cite{journals/corr/ChenZZL17} & 2.15 & 9.44 & 27.04 & 38.63 & 48.07 & 55.36 & 11.11 & 22.22 & 38.89 & 55.56 & \textbf{77.78} & 83.33 & 23.79 & 33.87 & 42.74 & 47.58 & 55.65 & 59.68 \\ 
\hline
JSTL-DGD \cite{conf/cvpr/XiaoLOW16} & 26.92 & 61.54 & 80.00 & 88.46 & 92.31 & 94.61 & 33.33 & 33.33 & 33.33 & 55.56 & 55.56 & 66.67 & 38.10 & 59.52 & 71.43 & 76.19 & 88.10 & 92.86 \\
ResNet-50 \cite{conf/cvpr/HeZRS16} & 34.62 & 80.00 & 93.85 & 96.92 & 98.46 & 99.23 & 44.44 & \textbf{55.56} & \textbf{55.56} & 55.56 & \textbf{77.78} & 77.78 & 61.90 & 73.81 & 90.48 & \textbf{95.24} & \textbf{95.24} & \textbf{95.24} \\
DCGAN+LSRO \cite{journals/corr/ZhengZY17} & 36.15 & 78.46 & 94.62 & 96.15 & \textbf{99.23} & 99.23 & 44.44 & \textbf{55.56} & \textbf{55.56} & 55.56 & 55.56 & 77.78 & 64.29 & 71.43 & 88.10 & 90.48 & 92.86 & \textbf{95.24} \\
DeepFool \cite{conf/cvpr/Moosavi-Dezfooli16} & 34.62 & 78.46 & 94.63 & 96.92 & 96.92 & 99.23 & 44.44 & \textbf{55.56} & \textbf{55.56} & 55.56 & 66.67 & 77.78 & 64.29 & 76.19 & 90.48 & \textbf{95.24} & \textbf{95.24} & \textbf{95.24} \\
\hline
\textbf{APN}  & \textbf{43.85} & \textbf{82.31} & \textbf{96.92} & \textbf{98.46} & \textbf{99.23} & \textbf{100}  & \textbf{55.56} & \textbf{55.56} & \textbf{55.56} & \textbf{66.67} & \textbf{77.78} & 77.78 & \textbf{66.67}  & \textbf{78.57} & \textbf{92.86} & \textbf{95.24} & \textbf{95.24} & \textbf{95.24} \\ \hline \hline
Evaluation    & \multicolumn{18}{c}{Individual Verification} \\ \hline
t-LRDC \cite{journals/pami/ZhengGX16} & 15.54 & 39.89 & 51.44 & 68.49 & 78.10 & 87.63 & 15.23 & 32.15 & 51.82 & 67.56 & 73.54 & 89.13 & 16.57 & 37.40 & 48.98  & 58.83  & 70.76 & 90.17 \\
XICE \cite{8011476} & 34.64 & 61.58 & 84.87 & 90.68 & 96.86 & 97.21 & 33.33 & 36.11 & 55.56 & 55.56 & 72.22 & 88.89 & 18.63 & 48.94 & 71.89 & 81.64 & 89.71 & 97.43 \\
GOG+XQDA \cite{conf/cvpr/LiaoHZL15}  & 10.49 & 30.60 & 51.83 & 62.77 & 77.29 & 86.14 & 25.00 & 55.56 & 88.89 & 91.67 & 97.22 & \textbf{100} & 33.93 & 45.73 & 64.93 & 77.85 & 87.40 & 91.99 \\
LOMO+XQDA \cite{conf/cvpr/LiaoHZL15}  & 25.32 & 59.10 & 81.98 & 86.96 & 92.96 & 94.84 & 5.56 & 36.11 & 80.56 & 88.89 & 88.89 & 88.89 & 40.72 & 57.79 & 77.78 & 86.90 & 94.44 & 96.03 \\
hiphop+CRAFT \cite{journals/corr/ChenZZL17} & 31.75 & 62.59 & 84.09 & 91.42 & 93.55 & 96.30 & \textbf{50.00} & \textbf{72.22} & \textbf{100} & \textbf{100} & \textbf{100} & \textbf{100} & 42.39 & 63.27 & 77.38 & 89.58 & 95.68 & \textbf{99.18} \\ 
\hline
JSTL-DGD \cite{conf/cvpr/XiaoLOW16} & 47.23 & 63.85 & 86.92 & 93.73 & 93.73 & 97.53 & 33.33 & 48.15 & 59.26 & 72.84 & 72.84 & 82.72 & 53.74 & 78.18 & 81.67 & 92.15 & 92.15 & 94.87 \\
ResNet-50 \cite{conf/cvpr/HeZRS16} & 82.26 & 95.86 & 98.54 & 99.38 & \textbf{99.58} & \textbf{99.58} & 44.44 & 50.00 & 72.22 & 77.78 & 83.33 & 83.33 & 76.19 & 91.67 & \textbf{95.24} & \textbf{95.24} & 95.24 & 95.24 \\ 
DCGAN+LSRO \cite{journals/corr/ZhengZY17} & 81.71 & 95.36 & 98.33 & 98.96 & 99.17 & \textbf{99.58} & 44.44 & 61.11 & 72.22 & 77.78 & 83.33 & 88.83 & 73.81 & 90.48 & \textbf{95.24} & \textbf{95.24} & 95.24 & 95.24 \\
DeepFool \cite{conf/cvpr/Moosavi-Dezfooli16} & 82.26 & 95.86 & 95.86 & 98.96 & 99.17 & \textbf{99.58} & 44.44 & 61.11 & 72.22 & 77.78 & 83.33 & 83.33 & 75.61 & 91.67 & \textbf{95.24} & \textbf{95.24} & 95.24 & 95.24 \\
\hline
\textbf{APN} & \textbf{84.00} & \textbf{96.72} & \textbf{98.69} & \textbf{99.58} & \textbf{99.58} & \textbf{99.58} & 44.44 & 61.11 & 77.78 & 77.78 & 83.33 & 88.89 & \textbf{79.54} & \textbf{94.05} & \textbf{95.24} & \textbf{95.24} & \textbf{97.15} & 97.15 \\
\hline
\end{tabular}}
\end{table}

\subsection{Comparison with Open-world Re-id Methods}
Open-world re-id is still under studied, and t-LRDC \cite{journals/pami/ZhengGX16} and XICE \cite{8011476} are two represented existing methods designed for the open-world setting in person re-id.
The results are reported in \tref{t:state-of-art}.
Our APN outperformed t-LRDC and XICE in almost all cases, and the margin is especially large on CUHK03.
Compared to t-LRDC and XICE, our APN is an end-to-end learning framework and takes adversarial learning into account for feature learning, so that APN is more tolerant to the attack of samples of non-target people.

\subsection{Comparison with Closed-world Re-id Methods}
We compared our method with related popular re-id methods developed for closed-world person re-identification. We mainly evaluated ResNet-50
\cite{conf/cvpr/HeZRS16}, XQDA \cite{conf/cvpr/LiaoHZL15}, CRAFT \cite{journals/corr/ChenZZL17}, DCGAN+LSRO \cite{journals/corr/ZhengZY17}, and
JSTL-DGD \cite{conf/cvpr/XiaoLOW16} for comparison.
These methods were all evaluated by following the same setting as our APN.
As shown in \tref{t:state-of-art}, these approaches optimal for closed-world scenario cannot well adapt to the open-world setting.
In the cases of Market-1501 and CUHK03, the proposed APN achieved overall better performance, especially when tested on Set Verification and when FTR is 0.1\% as compared to the others.
On Market-1501, APN obtained 7.7 more matching rate than the second place DCGAN+LSRO, which is also an application of GAN in re-id problem, when FTR is 0.1\% on Set Verification, and as well outperformed DCGAN+LSRO on all conditions on Individual Verification.
Although deep learning methods may cause overfitting problem on CUHK01 dataset, which is shown in Individual Verification of CUHK01 where hiphop+CRAFT had the best performance, APN outperformed all other deep learning methods on all conditions of CUHK01. When FTR is 0.1\% on CUHK01 Set Verification, APN gained 11.12 more matching rate as compared to DCGAN+LSRO and 5.56 matching rate more when FTR is 5\% on Individual Verification.
The compared closed-world models were designed with the assumption that the same identities hold between gallery and probe sets, while the relation between target and non-target people is not modelled. Meanwhile our APN is designed for the open-world group-based verification for discriminating target from non-target people.

\begin{table}[t]
\caption{Different generated imposter sources}
\label{t:imposter}
\parbox{\linewidth}{
\resizebox{\linewidth}{!}{
\begin{tabular}{c|c c c c c c|c c c c c c|c c c c c c}
\hline
Dataset & \multicolumn{6}{c|}{Market-1501} & \multicolumn{6}{c|}{CUHK01} & \multicolumn{6}{c}{CUHK03} \\ \hline
FTR  & 0.1\%            & 1\%              & 5\%              & 10\%             & 20\%             & 30\%  & 0.1\%      & 1\%    & 5\%              & 10\%             & 20\%             & 30\%   & 0.1\%      & 1\%    & 5\%              & 10\%             & 20\%             & 30\%        \\ \hline
Evaluation     & \multicolumn{18}{c}{Set Verification}                                                                                \\ \hline

APN         & \textbf{43.85} & \textbf{82.31} & \textbf{96.92} & \textbf{98.46} & \textbf{99.23} & \textbf{100}  & \textbf{55.56} & \textbf{55.56} & \textbf{55.56} & \textbf{66.67} & \textbf{77.78} & \textbf{77.78} & \textbf{66.67}  & \textbf{78.57} & \textbf{92.86} & \textbf{95.24} & \textbf{95.24} & \textbf{95.24} \\
APN w/o $D_t$   & 36.92  & 79.23          & 93.85          & 97.69          & \textbf{99.23}          & 99.23 & 44.44 & \textbf{55.56} & \textbf{55.56} & 55.56 & 55.56 & 66.67 &  61.90 & 73.81          & 88.10 & 90.48 & 92.86 & \textbf{95.24} \\
APN w/o $D_p$ & 35.38 & 78.46 & 93.85 & 95.38 & \textbf{99.23} & 99.23 & 44.44 & \textbf{55.56} & \textbf{55.56} & 55.56 & 55.56 & 66.67 & 61.90 & 71.43 & 88.10 & 92.86 & \textbf{95.24} & \textbf{95.24} \\
APN w/o $WS$ & 28.46 & 65.38 & 84.62 & 89.23 & 96.15 & 96.15 & 44.44 & \textbf{55.56} & \textbf{55.56} & 55.56 & 55.56 & 66.67 & 57.14 & 73.81 & 83.33 & 90.48 & 92.86 & 92.86 \\
No Imposter & 34.62 & 80.00 & 93.85 & 96.92 & 98.46 & 99.23 & 44.44 & \textbf{55.56} & \textbf{55.56} & 55.56 & \textbf{77.78} & \textbf{77.78} & 61.90 & 73.81 & 90.48 & \textbf{95.24} & \textbf{95.24} & \textbf{95.24} \\ \hline \hline
Evaluation     & \multicolumn{18}{c}{Individual Verification}                                                                                \\ \hline
APN & \textbf{84.00} & \textbf{96.72} & \textbf{98.69} & \textbf{99.58} & \textbf{99.58} & \textbf{99.58} & \textbf{44.44} & \textbf{61.11} & \textbf{77.78} & \textbf{77.78} & \textbf{83.33} & \textbf{88.89} & \textbf{79.54} & \textbf{94.05} & \textbf{95.24} & \textbf{95.24} & \textbf{97.15} & \textbf{97.15} \\
APN w/o $D_t$ & 82.92 & 95.40 & 98.13 & 98.75 & \textbf{99.58} & \textbf{99.58} & \textbf{44.44} & 55.56 & \textbf{77.78} & \textbf{77.78} & \textbf{83.33} & 83.33 & 73.81 & 90.48 & \textbf{95.24} & \textbf{95.24} & 95.24 & 95.24  \\
APN w/o $D_p$ & 82.26 & 95.06 & 98.13 & 98.75 & \textbf{99.58} & \textbf{99.58} & \textbf{44.44} & 50.00 & \textbf{77.78} & \textbf{77.78} & \textbf{83.33} & 83.33 & 72.62 & 90.48 & 91.67 & \textbf{95.24} & 95.24 & 95.24 \\
APN w/o $WS$ & 72.67 & 91.11 & 97.36 & 97.36 & 97.78 & 97.78 & \textbf{44.44} & \textbf{61.11} & \textbf{77.78
} & \textbf{77.78} & \textbf{83.33} & 83.33 & 69.05 & 84.52 & 94.05 & \textbf{95.24} & 95.24 & 95.24 \\
No Imposter & 82.26 & 95.86 & 98.54 & 99.38 & \textbf{99.58} & \textbf{99.58} & \textbf{44.44} & 50.00 & 72.22 & \textbf{77.78} & \textbf{83.33} & 83.33 & 76.19 & 91.67 & \textbf{95.24} & \textbf{95.24} & 95.24 & 95.24 \\ \hline
\end{tabular}}}
\end{table}

\subsection{Comparison with Related Adversarial Generation}
We compared our model with fine-tuned ResNet-50 with adversarial samples generated by DeepFool
\cite{conf/cvpr/Moosavi-Dezfooli16}, which is also a method using extra generated samples. DeepFool produced adversarial samples to fool the network by adding noises computed by gradients.
As shown in \tref{t:state-of-art}, our APN performed much better than DeepFool on all conditions. DeepFool cannot adapt to open-world re-id well because the adversarial samples generated are produced with a separate learning from the classifier learning and thus the relation between the generated samples and target set is not modelled for group-based verification, while in our APN we aim to generate target-like samples so as to make adversarial learning facilitate learning better features.

\begin{table}[t]
\captionsetup{font=scriptsize}
\parbox{.5\linewidth}{
\caption{Number of shots on Set Verification}
\label{t:one-shot_s}
\resizebox{\linewidth}{!}{
\begin{tabular}{c|c c c c c c | c c c c c c}
\hline
Method    & \multicolumn{6}{c|}{APN}                  & \multicolumn{6}{c}{ResNet-50}                                 \\ \hline
FTR         & 0.1\%          & 1\%   & 5\%   & 10\%  & 20\%  & 30\%  & 0.1\%          & 1\%            & 5\%            & 10\%           & 20\%           & 30\%           \\ \hline
Dataset &  \multicolumn{12}{|c}{Market-1501} \\ \hline
single-shot & \textbf{23.85} & \textbf{66.15} & \textbf{86.92} & \textbf{94.62} & \textbf{97.69} & \textbf{99.23} & 20.77 & 65.38 & 86.15 & 90.77 & 95.38 & 97.69 \\ \hline
multi-shot & \textbf{43.85} & \textbf{82.31} & \textbf{96.92} & \textbf{98.46} & \textbf{99.23} & \textbf{100} & 34.62 & 80.00 & 93.85 & 96.92 & 98.46 & 99.23 \\ \hline \hline
   Dataset  &  \multicolumn{12}{|c}{CUHK01} \\ \hline
single-shot & \textbf{33.33}  & \textbf{33.33} & \textbf{55.56} & \textbf{66.67} & \textbf{77.78} & \textbf{77.78} & \textbf{33.33} & \textbf{33.33} & 44.44 & 55.56 & 55.56  & \textbf{77.78} \\ \hline
multi-shot & \textbf{55.56} & \textbf{55.56} & \textbf{55.56} & \textbf{66.67} & \textbf{77.78} & \textbf{77.78} & 44.44 & \textbf{55.56} & \textbf{55.56} & 55.56 & \textbf{77.78} & \textbf{77.78} \\ \hline \hline
   Dataset  &  \multicolumn{12}{|c}{CUHK03} \\ \hline
single-shot & \textbf{59.52}  & \textbf{73.81} & \textbf{88.10} & \textbf{95.24} & \textbf{95.24} & \textbf{95.24} & 57.14     & 71.43          & \textbf{88.10}          & 90.48          & \textbf{95.24}          & \textbf{95.24}          \\ \hline
multi-shot & \textbf{66.67}  & \textbf{78.57} & \textbf{92.86} & \textbf{95.24} & \textbf{95.24} & \textbf{95.24}  & 61.90 & 73.81 & 90.48 & \textbf{95.24} & \textbf{95.24} & \textbf{95.24} \\ \hline
\end{tabular}}
}
\parbox{.5\linewidth}{
\caption{Number of shots on Individual Verification}
\label{t:one-shot_i}
\resizebox{\linewidth}{!}{
\begin{tabular}{c|c c c c c c|c c c c c c}
\hline
Method    & \multicolumn{6}{c|}{APN}                  & \multicolumn{6}{c}{ResNet-50}                                                          \\ \hline
FTR         & 0.1\%          & 1\%   & 5\%   & 10\%  & 20\%  & 30\%  & 0.1\%          & 1\%            & 5\%            & 10\%           & 20\%           & 30\%           \\ \hline
    Dataset &  \multicolumn{12}{c}{Market-1501} \\ \hline
single-shot & \textbf{80.04} & \textbf{98.08} & \textbf{99.17} & \textbf{100} & \textbf{100} & \textbf{100} & 80.00 & 95.03 & \textbf{99.17} & 99.17 & \textbf{100} & \textbf{100} \\
multi-shot & \textbf{84.00} & \textbf{96.72} & \textbf{98.69} & \textbf{99.58} & \textbf{99.58} & \textbf{99.58} & 82.26 & 95.86 & 98.54 & 99.38 & \textbf{99.58} & \textbf{99.58}  \\ \hline \hline
    Dataset  &  \multicolumn{12}{c}{CUHK01} \\ \hline
single-shot & \textbf{33.33}  & \textbf{66.67} & \textbf{88.89} & \textbf{88.89} & \textbf{88.89} & \textbf{88.89} & \textbf{33.33} & 55.56 & \textbf{88.89} & \textbf{88.89} & \textbf{88.89} & \textbf{88.89} \\
multi-shot & \textbf{44.44} & \textbf{61.11} & \textbf{77.78} & \textbf{77.78} & \textbf{83.33} & \textbf{88.89} & \textbf{44.44} & 50.00 & 72.22 & \textbf{77.78} & \textbf{83.33} & 83.33 \\ \hline \hline
    Dataset  &  \multicolumn{12}{c}{CUHK03} \\ \hline
single-shot & \textbf{78.57}  & \textbf{92.86} & \textbf{95.24} & \textbf{95.24} & \textbf{95.24} & \textbf{95.24} & 76.19 & 88.10 & \textbf{95.24} & \textbf{95.24} & \textbf{95.24} & \textbf{95.24} \\
multi-shot & \textbf{79.54} & \textbf{94.05} & \textbf{95.24} & \textbf{95.24} & \textbf{97.15} & \textbf{97.15}  & 76.19 & 91.67 & \textbf{95.24} & \textbf{95.24} & 95.24 & 95.24 \\ \hline
\end{tabular}}
}
\end{table}

\pgfplotsset{tiny, width=0.5\linewidth}

\begin{table}[t]
\captionsetup{font=scriptsize}
\parbox{.5\linewidth}{
\caption{Different target proportion of Market-1501 on Set Verification (TP. stands for Target Proportion)}
\label{t:dtpm}
\resizebox{\linewidth}{!}{
\begin{tabular}{c|c c c c c c|c c c c c c}
\hline
Method & \multicolumn{6}{c|}{APN} & \multicolumn{6}{c}{ResNet-50} \\ \hline
FTR         & 0.1\%            & 1\%              & 5\%              & 10\%             & 20\%             & 30\% & 0.1\%            & 1\%              & 5\%              & 10\%             & 20\%             & 30\% \\ \hline
TP. 0.5\% & \textbf{95.08} & \textbf{100} & \textbf{100} & \textbf{100} & \textbf{100} & \textbf{100} & 91.80 & \textbf{100} & \textbf{100} & \textbf{100} & \textbf{100} & \textbf{100} \\
TP. 1\% & \textbf{43.85} & \textbf{82.31} & \textbf{96.92} & \textbf{98.46} & \textbf{99.23} & \textbf{100} & 34.62 & 80.00 & 93.85 & 96.92 & 98.46 & 99.23 \\
TP. 3\% & \textbf{41.95} & \textbf{70.98} & \textbf{86.02} & \textbf{91.56} & \textbf{95.51} & \textbf{96.31} & 37.20 & 63.32 & 84.96 & 90.77 & 93.93 & 95.51 \\
TP. 5\% & \textbf{38.82} & \textbf{65.23} & \textbf{85.86} & \textbf{91.78} & \textbf{96.38} & \textbf{97.70} & 35.53 & 63.32 & 82.89 & 89.97 & 94.74 & 96.86 \\ \hline
\end{tabular}}
}
\parbox{.5\linewidth}{
\caption{LSRI vs. LSRO}
\label{t:gt}
\resizebox{\linewidth}{!}{
\begin{tabular}{c|c c c c c c|c c c c c c}
\hline
Evaluation & \multicolumn{6}{c|}{Set Verification} & \multicolumn{6}{c}{Individual Verification} \\ \hline
FTR         & 0.1\%            & 1\%              & 5\%              & 10\%             & 20\%             & 30\%      & 0.1\%            & 1\%              & 5\%              & 10\%             & 20\%             & 30\%         \\ \hline
Dataset     & \multicolumn{12}{c}{Market-1501}                                                                                \\ \hline
LSRI         & \textbf{43.85} & \textbf{82.31} & \textbf{96.92} & \textbf{98.46} & \textbf{99.23} & \textbf{100} & \textbf{84.00} & \textbf{96.72} & \textbf{98.69} & \textbf{99.58} & \textbf{99.58} & \textbf{99.58} \\
LSRO \cite{journals/corr/ZhengZY17}        & 40.00           & 80.00          & 93.85        & 96.92          & 98.46          & 99.23  & 82.00         & 95.36       & 98.33          & 98.75          & \textbf{99.58}          & \textbf{99.58} \\ \hline \hline
Dataset     & \multicolumn{12}{c}{CUHK01}                                                                                     \\ \hline
LSRI    & \textbf{55.56} & \textbf{55.56} & \textbf{55.56} & \textbf{66.67} & \textbf{77.78} & \textbf{77.78}   & \textbf{44.44} & \textbf{61.11} & \textbf{77.78} & \textbf{77.78} & \textbf{83.33} & \textbf{88.89}  \\
LSRO \cite{journals/corr/ZhengZY17}        & 44.44 & \textbf{55.56}    & \textbf{55.56}          & 55.56 & 55.56          & 66.67    & \textbf{44.44} & 55.56   & \textbf{77.78}     & \textbf{77.78} & \textbf{83.33}  & 83.33       \\ \hline \hline
Dataset     & \multicolumn{12}{c}{CUHK03}                                                                                     \\ \hline
LSRI    & \textbf{66.67}  & \textbf{78.57} & \textbf{92.86} & \textbf{95.24} & \textbf{95.24} & \textbf{95.24} & \textbf{79.54} & \textbf{94.05} & \textbf{95.24} & \textbf{95.24} & \textbf{97.15} & \textbf{97.15} \\
LSRO \cite{journals/corr/ZhengZY17}        & 64.29          & 73.81          & 88.10          & 90.48          & 92.86          & \textbf{95.24}    & 75.00 & 91.67 & \textbf{95.24} & \textbf{95.24} & 95.24 & 95.24      \\ \hline
\end{tabular}}
}
\end{table}

\subsection{Further Evaluation of Our Method}

\begin{description}[style=unboxed,leftmargin=0cm]
    \item[Effect of Generated Imposters.]
    It can be observed that, training with the imposters generated by APN can achieve large improvement as compared to the case without it, because these imposters are target-like and can improve the discriminating ability of the features.  The results of baseline without any generated imposters are shown in the rows indicated by “No Imposters” in \tref{t:imposter}.
    In details, on Set Verification, APN outperformed 9.23 matching rate on Market-1501, 11.12 more matching rate on CUHK01 and 4.77 on CUHK03 when FTR is 0.1\%.
    On Individual Verification, APN also has better performance on all cases.

    \item[Effect of Weight Sharing.]
    The weight sharing between the target discriminator and the feature extractor aims to ensure that the generator can learn from the feature extractor and generate more target-like attack samples. 
    Without the sharing, there is no connection between generation and feature extraction. 
    Taking Set Verification on Market for instance, ours degrades from 82.31\% to 65.38\% (no sharing，indicated by ``APN w/o WS'') when FTR=1\% in \tref{t:imposter}.

    \item[Effect of Person Discriminator and Target Discriminator.]
    Our APN is based on GAN consisting of generator, person discriminator $D_p$ and target discriminator $D_t$.
    To further evaluate them, we compared with APN without person discriminator (APN w/o $D_p$) and APN without target discriminator (APN w/o $D_t$).
    APN without target discriminator can be regarded as two independent components DCGAN and feature extraction network.
    To fairly compare these cases, LSRI was also applied as in APN for the generated samples.
    The results are reported in \tref{t:imposter}.
    It is obvious that our full APN is the most effective one among the compared cases.
    Sometimes generating imposters by APN without person discriminator $D_p$ or target discriminator $D_t$ even degrade the performance as compared to the case of no imposter.
    When target discriminator is discarded, although person-like images can be generated, they are not similar to target people
    and thus are not serious attacks to the features for group-based verification.
    In the case without person discriminator, the generator even fails to generate person-like images (see \fref{fig:samples}) so that the performance is also degraded in most cases.
    This indicates that the person discriminator plays an important role in generating person-like images, and the target discriminator is significant for helping the generator to generate better target-like imposters, so that the feature extractor can benefit more from distinguishing these imposters.

    \item[LSRI vs LSRO.] We verified that the modification of LSRO, namely LSRI in \qref{eq:LSRI} is more suitable for optimizing the open-world re-id. The performance of comparing our LSRI with the original LSRO is reported in \tref{t:gt}. It shows that
       the feature extractor is more likely to correctly discriminate target people under the same FTR using LSRI on. It is proved that our modification LSRI is more appropriate for open-world re-id scenario, since the imposters are allocated equal probabilities of being
        non-target for group-based towards modelling, so they are more likely to be far away from target person samples, leading to more discriminative feature representation for target people, while in LSRO, the imposters are allocated equal probabilities of being non-target as well as target.

   \item[Effect of target proportion.] The evaluation results on different target proportion are reported in \tref{t:dtpm}. We used different percentages of people marked as target.
        This verification was conducted on Market-1501, and we used original ResNet-50 for comparison. While TTR declines with the growth of target proportion due to more target people to verify, our APN can still outperformed the original ResNet-50 in all cases.

    \item[Effect of the Number of Shots.]
    The performance under multi-shot and single-shot settings were also compared in our experiments. For multi-shot setting, we randomly selected two images of each target person as gallery set, while for single-shot setting, we only selected one. As shown in \tref{t:one-shot_s} and \tref{t:one-shot_i}, on both single-shot and multi-shot settings, our APN outperformed
ResNet-50 on all conditions of Market-1501, CUHK01, and CUHK03. Especially on Set Verification, for CUHK01, when FTR is 10\%, APN outperformed ResNet-50 11.11\% under both single-shot and multi-shot settings.
\end{description}

\section{Conclusion}

For the first time, we demonstrate how adversarial learning can be used to solve the open-world group-based person re-id problem. The introduced adversarial person re-id enables a mutually related and cooperative progress among learning to represent, learning to generate, learning to attack, and learning to defend. In addition, this adversarial modelling is also further improved by a label smoothing regularization for imposters under semi-supervised learning.

\section*{Acknowledgment}
This work was supported partially by the National Key Research and Development Program of China (2016YFB1001002) and the NSFC(61522115, 61472456), Guangdong Programme (2016TX03X157), and the Royal Society Newton Advanced Fellowship (NA150459). The corresponding author for this paper is Wei-Shi Zheng.

\clearpage

\bibliographystyle{splncs04}
\bibliography{reference}

\begin{thebibliography}{10}
\providecommand{\url}[1]{\texttt{#1}}
\providecommand{\urlprefix}{URL }
\providecommand{\doi}[1]{https://doi.org/#1}

\bibitem{conf/cvpr/AhmedJM15}
Ahmed, E., Jones, M.J., Marks, T.K.: An improved deep learning architecture for
  person re-identification. In: CVPR. IEEE Computer Society (2015)

\bibitem{baltieri2011_308}
Baltieri, D., Vezzani, R., Cucchiara, R.: 3dpes: 3d people dataset for
  surveillance and forensics. In: Proceedings of the 1st International ACM
  Workshop on Multimedia access to 3D Human Objects. pp. 59--64. Scottsdale,
  Arizona, USA (Nov 2011)

\bibitem{journals/ivc/Bedagkar-GalaS14}
Bedagkar-Gala, A., Shah, S.K.: A survey of approaches and trends in person
  re-identification. Image Vision Comput.  \textbf{32}(4),  270--286 (2014)

\bibitem{conf/bmvc/CancelaHG14}
Cancela, B., Hospedales, T.M., Gong, S.: Open-world person re-identification by
  multi-label assignment inference. In: Valstar, M.F., French, A.P., Pridmore,
  T.P. (eds.) BMVC. BMVA Press (2014)

\bibitem{journals/tcsv/ChenZLY17}
Chen, Y.C., Zheng, W.S., Lai, J.H., Yuen, P.C.: An asymmetric distance model
  for cross-view feature mapping in person reidentification. IEEE Trans.
  Circuits Syst. Video Techn.  \textbf{27}(8),  1661--1675 (2017)

\bibitem{journals/corr/ChenZZL17}
Chen, Y.C., Zhu, X., Zheng, W.S., Lai, J.H.: Person re-identification by camera
  correlation aware feature augmentation. CoRR  \textbf{abs/1703.08837} (2017)

\bibitem{conf/cvpr/ChengGZWZ16}
Cheng, D., Gong, Y., Zhou, S., Wang, J., Zheng, N.: Person re-identification by
  multi-channel parts-based cnn with improved triplet loss function. In: CVPR.
  IEEE Computer Society (2016)

\bibitem{deng2018image}
Deng, W., Zheng, L., Ye, Q., Kang, G., Yang, Y., Jiao, J.: Image-image domain
  adaptation with preserved self-similarity and domain-dissimilarity for person
  reidentification. In: CVPR. p.~6. IEEE Computer Society (2018)

\bibitem{journals/corr/DingLWC15}
Ding, S., Lin, L., Wang, G., Chao, H.: Deep feature learning with relative
  distance comparison for person re-identification. CoRR
  \textbf{abs/1512.03622} (2015)

\bibitem{conf/cvpr/FarenzenaBPMC10}
Farenzena, M., Bazzani, L., Perina, A., Murino, V., Cristani, M.: Person
  re-identification by symmetry-driven accumulation of local features. In:
  CVPR. IEEE Computer Society (2010)

\bibitem{series/acvpr/978-1-4471-6295-7}
Gong, S., Cristani, M., Yan, S., Loy, C.C. (eds.): Person Re-Identification.
  Advances in Computer Vision and Pattern Recognition, Springer (2014)

\bibitem{conf/cvpr/HeZRS16}
He, K., Zhang, X., Ren, S., Sun, J.: Deep residual learning for image
  recognition. In: CVPR. IEEE Computer Society (2016)

\bibitem{conf/scia/HirzerBRB11}
Hirzer, M., Beleznai, C., Roth, P.M., Bischof, H.: Person re-identification by
  descriptive and discriminative classification. In: SCIA. Springer (2011)

\bibitem{conf/eccv/HirzerRKB12}
Hirzer, M., Roth, P.M., Köstinger, M., Bischof, H.: Relaxed pairwise learned
  metric for person re-identification. In: ECCV. Lecture Notes in Computer
  Science (2012)

\bibitem{Kawanishi_shinpuhkan2014:a}
Kawanishi, Y., Wu, Y., Mukunoki, M., Minoh, M.: Shinpuhkan2014: A multi-camera
  pedestrian dataset for tracking people across multiple cameras

\bibitem{journals/corr/KingmaB14}
Kingma, D.P., Ba, J.: Adam: A method for stochastic optimization. CoRR
  \textbf{abs/1412.6980} (2014)

\bibitem{journals/pami/KviatkovskyAR13}
Kviatkovsky, I., Adam, A., Rivlin, E.: Color invariants for person
  reidentification. IEEE Trans. Pattern Anal. Mach. Intell.  \textbf{35}(7),
  1622--1634 (2013)

\bibitem{conf/cvpr/KostingerHWRB12}
Köstinger, M., Hirzer, M., Wohlhart, P., Roth, P.M., Bischof, H.: Large scale
  metric learning from equivalence constraints. In: CVPR. IEEE Computer Society
  (2012)

\bibitem{li2012human}
Li, W., Zhao, R., Wang, X.: Human reidentification with transferred metric
  learning. In: ACCV (2012)

\bibitem{conf/cvpr/LiZXW14}
Li, W., Zhao, R., Xiao, T., Wang, X.: Deepreid: Deep filter pairing neural
  network for person re-identification. In: CVPR. IEEE Computer Society (2014)

\bibitem{li2014deepreid}
Li, W., Zhao, R., Xiao, T., Wang, X.: Deepreid: Deep filter pairing neural
  network for person re-identification. In: CVPR (2014)

\bibitem{conf/cvpr/LiaoHZL15}
Liao, S., Hu, Y., Zhu, X., Li, S.Z.: Person re-identification by local maximal
  occurrence representation and metric learning. In: CVPR. pp. 2197--2206. IEEE
  Computer Society (2015)

\bibitem{journals/pr/LiuGL14}
Liu, C., Gong, S., Loy, C.C.: On-the-fly feature importance mining for person
  re-identification. Pattern Recognition  \textbf{47}(4),  1602--1615 (2014)

\bibitem{journals/corr/LuIF17}
Lu, J., Issaranon, T., Forsyth, D.A.: Safetynet: Detecting and rejecting
  adversarial examples robustly. CoRR  \textbf{abs/1704.00103} (2017)

\bibitem{ma2012bicov}
Ma, B., Su, Y., Jurie, F.: Bicov: a novel image representation for person
  re-identification and face verification. In: BMVC (2012)

\bibitem{DBLP:journals/ivc/MaSJ14}
Ma, B., Su, Y., Jurie, F.: Covariance descriptor based on bio-inspired features
  for person re-identification and face verification. Image Vision Comput.
  \textbf{32}(6-7),  379--390 (2014). \doi{10.1016/j.imavis.2014.04.002}

\bibitem{conf/cvpr/MignonJ12}
Mignon, A., Jurie, F.: Pcca: A new approach for distance learning from sparse
  pairwise constraints. In: CVPR. IEEE Computer Society (2012)

\bibitem{conf/cvpr/Moosavi-Dezfooli16}
Moosavi-Dezfooli, S.M., Fawzi, A., Frossard, P.: Deepfool: A simple and
  accurate method to fool deep neural networks. In: CVPR. IEEE Computer Society
  (2016)

\bibitem{journals/corr/PapernotMJFCS15}
Papernot, N., McDaniel, P.D., Jha, S., Fredrikson, M., Celik, Z.B., Swami, A.:
  The limitations of deep learning in adversarial settings. CoRR
  \textbf{abs/1511.07528} (2015)

\bibitem{journals/corr/RadfordMC15}
Radford, A., Metz, L., Chintala, S.: Unsupervised representation learning with
  deep convolutional generative adversarial networks. CoRR
  \textbf{abs/1511.06434} (2015)

\bibitem{conf/nips/SubramaniamCM16}
Subramaniam, A., Chatterjee, M., Mittal, A.: Deep neural networks with inexact
  matching for person re-identification. In: Lee, D.D., Sugiyama, M., von
  Luxburg, U., Guyon, I., Garnett, R. (eds.) NIPS (2016)

\bibitem{journals/corr/SzegedyZSBEGF13}
Szegedy, C., Zaremba, W., Sutskever, I., Bruna, J., Erhan, D., Goodfellow,
  I.J., Fergus, R.: Intriguing properties of neural networks. CoRR
  \textbf{abs/1312.6199} (2013)

\bibitem{journals/tcsv/TaoJWYL13}
Tao, D., Jin, L., Wang, Y., Yuan, Y., Li, X.: Person re-identification by
  regularized smoothing kiss metric learning. IEEE Trans. Circuits Syst. Video
  Techn.  \textbf{23}(10),  1675--1685 (2013)

\bibitem{conf/icip/WangZXG16}
Wang, H., Zhu, X., Xiang, T., Gong, S.: Towards unsupervised open-set person
  re-identification. In: ICIP. IEEE (2016)

\bibitem{journals/corr/WangGZW16}
Wang, T., Gong, S., Zhu, X., Wang, S.: Person re-identification by
  discriminative selection in video ranking. IEEE transactions on pattern
  analysis and machine intelligence  \textbf{38}(12),  2501--2514 (2016)

\bibitem{journals/corr/WuSH16}
Wu, L., Shen, C., van~den Hengel, A.: Personnet: Person re-identification with
  deep convolutional neural networks. CoRR  \textbf{abs/1601.07255} (2016)

\bibitem{journals/corr/WuCLWYZ16}
Wu, S., Chen, Y.C., Li, X., Wu, A., You, J., Zheng, W.S.: An enhanced deep
  feature representation for person re-identification. CoRR
  \textbf{abs/1604.07807} (2016)

\bibitem{conf/cvpr/XiaoLOW16}
Xiao, T., Li, H., Ouyang, W., Wang, X.: Learning deep feature representations
  with domain guided dropout for person re-identification. In: CVPR. IEEE
  Computer Society (2016)

\bibitem{zheng2015scalable}
Zheng, L., Shen, L., Tian, L., Wang, S., Wang, J., Tian, Q.: Scalable person
  re-identification: A benchmark. In: ICCV (2015)

\bibitem{conf/cvpr/ZhengGX11}
Zheng, W.S., Gong, S., Xiang, T.: Person re-identification by probabilistic
  relative distance comparison. In: CVPR. IEEE Computer Society (2011)

\bibitem{conf/cvpr/ZhengGX12}
Zheng, W.S., Gong, S., Xiang, T.: Transfer re-identification: From person to
  set-based verification. In: CVPR. IEEE Computer Society (2012)

\bibitem{journals/pami/ZhengGX16}
Zheng, W.S., Gong, S., Xiang, T.: Towards open-world person re-identification
  by one-shot group-based verification. IEEE Trans. Pattern Anal. Mach. Intell.
   \textbf{38}(3),  591--606 (2016)

\bibitem{journals/corr/ZhengZY17}
Zheng, Z., Zheng, L., Yang, Y.: Unlabeled samples generated by gan improve the
  person re-identification baseline in vitro. CoRR  \textbf{abs/1701.07717}
  (2017)

\bibitem{zhong2018camera}
Zhong, Z., Zheng, L., Zheng, Z., Li, S., Yang, Y.: Camera style adaptation for
  person re-identification. In: CVPR. pp. 5157--5166. IEEE Computer Society
  (2018)

\bibitem{journals/corr/ZhuZLLCZ17}
Zhu, J., Zeng, H., Liao, S., Lei, Z., Cai, C., Zheng, L.: Deep hybrid
  similarity learning for person re-identification. CoRR
  \textbf{abs/1702.04858} (2017)

\bibitem{8011476}
Zhu, X., Wu, B., Huang, D., Zheng, W.S.: Fast open-world person
  re-identification. IEEE Transactions on Image Processing  \textbf{PP}(99),
  ~1--1 (2017). \doi{10.1109/TIP.2017.2740564}

\end{thebibliography}
\end{document}